\documentclass[10pt,twocolumn,letterpaper]{article}

\usepackage[pagenumbers]{iccv} 

\usepackage{times}
\usepackage{float}
\usepackage{epsfig}
\usepackage{graphicx}
\usepackage{amsmath}
\usepackage{bm}
\usepackage{amssymb}
\usepackage{booktabs}
\usepackage{subcaption}
\usepackage{makecell}
\usepackage{rotating}
\usepackage{multirow}
\usepackage{cuted}
\usepackage{colortbl}
\usepackage{tcolorbox}
\usepackage{array}
\usepackage{enumitem}
\usepackage{multicol}
\usepackage[skip=2pt]{caption}
\usepackage{pifont}
\usepackage[accsupp]{axessibility}

\newcommand{\mAP}{\mathrm{mAP}}

\definecolor{dilcolor}{HTML}{d6466d}
\definecolor{ecdilcolor}{HTML}{5900b1}
\definecolor{cilcolor}{HTML}{00b159}
\definecolor{classcolor}{HTML}{00aedb}
\definecolor{objdetcolor}{HTML}{f37735}
\definecolor{ecrico}{HTML}{c5e0b4}
\definecolor{drico}{HTML}{bdd7ee}
\definecolor{keytakeaways}{HTML}{ffeecb}

\definecolor{iccvblue}{rgb}{0.21,0.49,0.74}
\usepackage[pagebackref,breaklinks,colorlinks,allcolors=iccvblue]{hyperref}

\def\paperID{} 
\def\confName{ICCV}
\def\confYear{2025}

\title{Incremental Object Detection with Prompt-based Methods}

\author{%
  Matthias~Neuwirth\textendash Trapp$^{1, 2}$\quad
  Maarten~Bieshaar$^{2}$\quad
  Danda~Pani~Paudel$^{3}$\quad
  Luc~Van~Gool$^{3}$\\
  {\small\texttt{mneuwirth@ethz.ch}}\\[0.6em]
  {\small$^{1}$ETH Zürich\quad}
  {\small$^{2}$Bosch Research\quad}
  {\small$^{3}$INSAIT, Sofia University “St.~Kliment Ohridski”}
}

\begin{document}
\maketitle
\begin{abstract}
Visual prompt-based methods have seen growing interest in incremental learning (IL) for image classification. These approaches learn additional embedding vectors while keeping the model frozen, making them efficient to train. However, no prior work has applied such methods to incremental object detection (IOD), leaving their generalizability unclear. In this paper, we analyze three different prompt-based methods under a complex domain-incremental learning setting. We additionally provide a wide range of reference baselines for comparison. Empirically, we show that the prompt-based approaches we tested underperform in this setting. However, a strong yet practical method—combining visual prompts with replaying a small portion of previous data—achieves the best results. Together with additional experiments on prompt length and initialization, our findings offer valuable insights for advancing prompt-based IL in IOD.
\end{abstract}
    
\section{Introduction}

In incremental learning (IL), models are sequentially trained on new tasks~\cite{wang_comprehensive_2024}. This work addresses domain incremental learning (DIL) for object detection, wherein each new task introduces data from a previously unseen domain, though target classes remain consistent across tasks~\cite{wang_comprehensive_2024, menezes_continual_2022}. When training on a new domain, the optimization process updates model weights to minimize task-specific losses, inadvertently overwriting previously learned representations. This phenomenon, termed \textit{catastrophic forgetting}, remains a central challenge in IL. Attempts to mitigate forgetting often reduce model adaptability to new tasks, resulting in the \textit{stability-plasticity dilemma}~\cite{kim_stability-plasticity_2023}.

Various strategies have been proposed to manage this dilemma, with one promising direction involving learnable prompts~\cite{zhou_continual_2024}. In prompt-based methods, trainable prompts are prepended to inputs of pre-trained transformers to guide task-specific feature extraction~\cite{zhou_continual_2024, wang_learning_2022, wang_dualprompt_2022}. Typically, distinct prompts are allocated per task, and during inference, the appropriate prompt is selected based on task identification~\cite{wang_learning_2022, wang_dualprompt_2022, wang_s-prompts_2022}. Visual prompts differ from textual ones, as they do not convey language meaning and can be considered \textit{pseudo-words}. Despite numerous variations, prompt-based methods have mostly been evaluated only on classification tasks, leaving their effectiveness in other computer vision objectives largely unexplored~\cite{zhou_continual_2024, menezes_continual_2022}. 

In this paper, we present the first study of prompt-based IL methods applied to DIL for object detection. We establish several robust baselines and systematically evaluate three widely-used prompt-based IL methods—L2P~\cite{wang_learning_2022}, DualPrompt~\cite{wang_dualprompt_2022}, and S-Prompt~\cite{wang_s-prompts_2022}—under varying configurations. We extend our analysis by examining prompt length and prompt initialization strategies.

Our experiments leverage the challenging D-RICO benchmark~\cite{anonymous_rico_2025}, consisting of 15 tasks from automotive and surveillance domains—key application areas for object detection. D-RICO integrates data from 14 diverse datasets, spanning imaging sensors, lens types, perspectives, environmental conditions (\eg, weather, daytime), and both synthetic and real-world scenarios. This benchmark thus embodies significant distributional shifts, offering a rigorous framework for evaluating IL methods.

We demonstrate that although the three examined prompt-based methods perform well on classification tasks, they significantly underperform on object detection within D-RICO. Combining these findings with detailed analyses of prompt initialization strategies and optimal prompt lengths, we provide a comprehensive understanding of the factors influencing prompt-based IL performance, thereby paving the way for future developments in this area.

Our main contributions are:
\begin{itemize}
    \item We are the first to study prompt-based IL for object detection, showing common methods underperform, with DualPrompt as the most effective tested method.
    \item Our work presents strong baselines and shows that combining visual prompt tuning with replaying previous task data is a practical and straightforward approach to IL.
    \item Further investigations show that choosing a fixed prompt length is sufficient across tasks, and initializing prompts with lower values is more effective. 
\end{itemize}

\section{Related Works}\label{sec:rel-work}

\subsection{Incremental Object Detection}
Object detection models broadly fall into two categories: single-stage detectors, which focus on fast inference~\cite{hussain_yolov1_2024, carion_end--end_2020}, and two-stage detectors, known for their higher accuracy~\cite{ren_faster_2017, cai_cascade_2021, fang_eva-02_2023, du_overview_2020}. In incremental learning (IL) for object detection, two-stage models have traditionally dominated research~\cite{leonardis_bridge_2025, wagner_forgetting_2024, song_non-exemplar_2024, menezes_continual_2022}, though incremental learning with single-stage is increasingly explored~\cite{mo_multi-level_2025, shieh_continual_2020, dong_incremental-detr_2023, liu_continual_2023}. To mitigate catastrophic forgetting, distillation-based regularization techniques~\cite{leonardis_bridge_2025, mo_multi-level_2025, dong_towards_2024, kang_alleviating_2023, dong_incremental-detr_2023, dong_class-incremental_2023, qian_contrastive_2022, gang_predictive_2022}, as well as rehearsal methods that replay previously seen data~\cite{kim_sddgr_2024, monte_replay_2024, yang_one-shot_2023, liu_augmented_2023, shieh_continual_2020}, have emerged as leading approaches. Additionally, representation-based strategies~\cite{mo_multi-level_2025, lu_few-shot_2024}, optimization-oriented methods~\cite{joseph_incremental_2022, li_raise_2022, wang_opennet_2023}, and various hybrid or novel methods~\cite{yang_continual_2022, liu_continual_2023, ibrahim_node_2024} are progressively expanding the scope of incremental object detection research. Nevertheless, compared to the extensive body of work on incremental classification, incremental learning for object detection remains relatively understudied~\cite{wang_comprehensive_2024}.

\subsection{Prompt-based Incremental Learning} 
\textit{Visual prompts} are a parameter-efficient fine-tuning technique to adapt pre-trained models to new data~\cite{jia_visual_2022}. The initial method L2P~\cite{wang_learning_2022} demonstrated the feasibility of applying visual prompts to IL. They learned a pool of these visual prompts and a corresponding key for each prompt. The visual prompts are selected using cosine similarity between the classification token and this key.
Following methods improve on this by distinguishing between general and expert prompts~\cite{wang_dualprompt_2022}, employing non-shared prompt pools~\cite{wang_s-prompts_2022}, prompt-selection through k-nearest neightbor search~\cite{wang_s-prompts_2022}, attention-based prompt combination~\cite{smith_coda-prompt_2023}, separate learning objectives~\cite{wang_hierarchical_2024}, or generating prompts using meta-networks~\cite{jung_generating_2023, yang_generating_2024, lu_training_2025}.
A further overview is provided by Wang et al.~\cite{wang_comprehensive_2024} and Zhou et al.~\cite{zhou_continual_2024}.
However, these prompt-based methods are not evaluated on other IL computer vision types than classification.
\section{Preliminary}

\subsection{Domain Incremental Object Detection}

We study the problem of domain incremental object detection, where a model is exposed to a sequence of tasks, \ie domains~\cite{wang_comprehensive_2024, zhang_few-shot_2025, wang_s-prompts_2022}. At step $t$, the model learns task $\mathcal{T}_t$ using the dataset $\mathcal{D}_t = (\mathcal{X}_t, \mathcal{Y}_t)$, where the image set $\mathcal{X}_t = \{\mathbf{x}_i^t\}_{i=1}^{n_t}$ consists of $n_t$ images. Each image $\mathbf{x}_i^t$ has dimensions $\mathbf{x}_i^t \in \mathbb{R}^{H_i^t \times W_i^t \times C_i^t}$, with $H_i^t$, $W_i^t$, and $C_i^t$ denoting height, width, and channel count, respectively.

The annotation set $\mathcal{Y}_t = \{\mathbf{y}_i^t\}_{i=1}^{n_t}$ corresponds to these images, where each $\mathbf{y}_i^t$ is a collection of object instances: $\mathbf{y}_i^t = \{(c_{i,j}^t, \mathbf{b}_{i,j}^t)\}_{j=1}^{m_i^t}$. Here, $c_{i,j}^t \in \mathcal{C}$ represents the class label of the $j$-th object in image $\mathbf{x}_i^t$, with $\mathcal{C}$ being the category set, which is fixed for this domain IOD setting, and $\mathbf{b}_{i,j}^t \in \mathbb{R}^4$ denotes the bounding box coordinates. The number of annotated objects $m_i^t$ may vary across images.

During training, the model has access to the task identity, but this information is not provided at test time. A model trained on task $\mathcal{T}_t$ using the data $\mathcal{D}_t$ is denoted by $\mathcal{M}_t$.

\subsection{Visual Prompt Tuning}

A visual prompt is a set of learnable parameters $\mathbf{p}\in\mathbb{R}^{L_p\times D}$, where $L_p$ is the prompt length, \ie the number of prompts, and $D$ is the embedding dimension~\cite{jia_visual_2022, wang_dualprompt_2022}. The backbone itself is kept frozen, and the visual prompts are incorporated into it and optimized during training. There are two prominent ways to incorporate the visual prompts into the backbone.

\begin{itemize}
    \item \textbf{Prompt Tuning (Pro-T).} The prompts are prepended to the key $h_K$, query $h_Q$ and value $h_V$ of the multi-head self-attention (MSA) layer.
    \begin{equation}
        f_\mathrm{prompt}^\mathrm{Pro-T}= \mathrm{MSA}([p;h_Q], [p;h_K], [p;h_V])
    \end{equation}
    Here, $[\cdot;\cdot]$ is the concatenation operation along the sequence length.
    The output sequence, compared to the non-prompted MSA, is extended by the length of the prompt.

    \item \textbf{Prefix Tuning (Pre-T).} The prompt is split into two parts that are prepended to the key and value, \ie $p_K,p_V\in\mathbb{R}^{L_p/2 \times D}$.
    \begin{equation}
        f_\mathrm{prompt}^\mathrm{Pre-T}= \mathrm{MSA}(h_Q, [p_K;h_K], [p_V;h_V])
    \end{equation}
    The length of the output sequence remains unchanged by the visual prompts.
\end{itemize}

\noindent More details can be found here~\cite{jia_visual_2022, wang_dualprompt_2022}.

\begin{table*}[htbp]
\centering
\footnotesize
\caption{Results for different prompting techniques and prompt-based IL methods on the D-RICO benchmark. Joint and individual training represent the upper bounds, Naïve FT the lower bound, and the two replay configurations are strong baselines. The three prompt-based IL methods fall behind even 1\% replay. Best IL approach in bold.}
\label{tabel:main-results}
\begin{tabular}{llcllll} 
\toprule
& & & \multicolumn{4}{c}{\textbf{Domain RICO}} \\ 
\cmidrule(lr){4-7} 
  \textbf{Method}  & \textbf{Prompt style} & \textbf{Freeze Head after 1. Task} & \multicolumn{1}{c}{\textbf{$\overline{\bm{\mAP}}$ \textuparrow}} & \multicolumn{1}{c}{\textbf{$\mathbf{FM}$  \textdownarrow}}  & \multicolumn{1}{c}{\textbf{$\mathbf{FWT}$ \textuparrow}} & \multicolumn{1}{c}{\textbf{$\mathbf{IM}$ \textuparrow}}  \\ 
\midrule
 Joint Training               & No Prompt & \ding{55} & 25.45 & - &- & -  \\

                  & Shallow Prompt& \ding{55} & 26.39 & - &-  &-  \\

                  & Deep Prompt &\ding{55}  & 29.55  & - &- & - \\
\midrule
  Individual Training                & No Prompt& \ding{55}& 26.92 & - & - & -   \\

   & Shallow Prompt& \ding{55} & 28.98 & - & - & -    \\

             &  Deep Prompt& \ding{55}  & 33.12 & - & - &  -   \\

\midrule

 Naïve FT           & No Prompt&\ding{55} & 16.20 & 13.17 & -7.31 &  -2.81 \\

                   & Shallow Prompt& \ding{55} & 20.88 &10.38  & -4.25 & -0.36   \\
                  
                &  Deep Prompt& \ding{55}  &  21.98& 16.60 & 2.54 &  5.71   \\
                
                & No Prompt&\ding{51} & 23.49 & 0 & -11.32 &  -6.05  \\
    
                  & Shallow Prompt&\ding{51} & 23.23 & 2.54 & -9.19 & -4.27   \\
                  
                &  Deep Prompt& \ding{51}  & 22.89 & 14.53 & 1.66 &   4.96  \\

\midrule
Replay 1\%        

                & No Prompt & \ding{55}& 21.44 & 7.08 & -6.81 &  -2.44  \\
         
                  & Shallow Prompt & \ding{55} & 23.16 & 6.79 & -4.96 & -0.95   \\

                  &  Deep Prompt  & \ding{55} & 26.55 & 10.74 & 2.43 &  5.60 \\
                  
                  & Shallow Prompt & \ding{51} & 23.27 & 2.30 & -9.29&  -4.42 \\

                  &  Deep Prompt  & \ding{51} & 26.94 & 9.71 & 1.77 &  5.16 \\

\midrule
Replay 10\%

                & No Prompt & \ding{55} & 25.41 & 2.81 & -6.11 & -1.88  \\

                  & Shallow Prompt &\ding{55}  & 26.79 & 3.64 & -3.64 & 0.14   \\

                  &  Deep Prompt  & \ding{55} & \textbf{31.62} & 4.63 & \textbf{2.60} &  \textbf{5.76} \\
                  
                  & Shallow Prompt & \ding{51} & 24.41 & \textbf{0.89} & -9.29& -4.41   \\

                  &  Deep Prompt  & \ding{51} & 31.15 & 3.59 & 1.11 &  \textbf{5.76} \\

\midrule
L2P~\cite{wang_learning_2022}
                  &  & \ding{55} & 20.92 & 10.33 & -4.28 & -0.35  \\

                  &  &\ding{51}  & 23.28 & 1.89 & -9.80& -4.76  \\
\midrule
DualPrompt~\cite{wang_dualprompt_2022} 
                  &  & \ding{55} & 18.61 & 12.29 & -5.16& -1.11   \\

                  &  & \ding{51} & 23.81  & 1.07 & -9.94& -4.91  \\
\midrule
S-Prompt~\cite{wang_s-prompts_2022} 
                  &  & \ding{55} & 20.71 & 10.27 & -4.61 & -0.62  \\
                &  & \ding{51} & 22.78 & 1.36 & -10.86  & -5.66 \\

\bottomrule
\end{tabular}
\end{table*}

\section{Experiments}

\subsection{Setup}
\noindent\textbf{Model.} We use the EVA-02 vision transformer~
\cite{fang_eva-02_2023} in its \textit{big} configuration. We include the prompts in the positional embedding but exclude them from the rotary embedding. We repeat the prompts on the window partitioning layers by the number of windows. We use the COCO pre-trained weights and freeze the backbone, region proposal network, and head, leaving only the output layer trainable.

\noindent\textbf{Optimization.} We employ the AdamW optimizer with a learning rate of 0.001 and cosine learning rate decay. We train each task for 1,000 iterations and a batch size of 10.

\subsubsection{Methods}
We select three prominent methods for the evaluation: L2P~\cite{wang_learning_2022}, DualPrompt~\cite{wang_dualprompt_2022}, and S-prompt~\cite{wang_s-prompts_2022}. While these are not the state-of-the-art (SOTA), their simplicity allows for a clearer understanding of the problem and provides valuable insights. We consider two different configurations: fixing the head after the first task and continuing to train the head.

We compare these prompt-based IL methods to a wide variety of reference baselines~\cite{anonymous_rico_2025}:
\begin{itemize}
    \item \textbf{Joint Training} merges all tasks into a single training dataset and trains a single model on these. The test datasets are separate. 
    \item \textbf{Individual Training} trains and tests a separate model for each task.
    \item \textbf{Naïve finetuning} (FT) trains a single model sequentially on the tasks without any IL method.
    \item \textbf{Replay} keeps a portion (1\% and 10\% in this case) for the sequential tasks to train on new and some old data at the same time.
\end{itemize}

\noindent We consider different configurations for these reference baselines:
\begin{itemize}
    \item \textbf{Freeze Head after 1. Task} to reduce model plasticity.
    \item \textbf{No Prompt} uses the standard EVA-02 model without modifications.
    \item \textbf{Shallow Prompt} uses a trainable 50 prompts and prepends them to the image embeddings before the first attention block~\cite{jia_visual_2022}.
    \item \textbf{Deep Prompt} learns 100 prompts for each layer and prepends them to the image embeddings~\cite{jia_visual_2022}.
\end{itemize}

\noindent All settings employ prompt tuning, with prefix tuning being used only in DualPrompt.

\subsection{Benchmark}

We employ the D-RICO benchmark~\cite{anonymous_rico_2025} as it offers the most diverse domain distribution shifts. It consists of 14 different datasets from which 15 tasks are created. These datasets encompass various camera sensors (RGB, thermal, gated, and event), lenses, viewpoints, time of day, weather conditions, and both real and synthetic domains. The output distribution also varies in terms of bounding box location, aspect ratio, and class ratios. Additionally, due to the origin of multiple datasets, the label quality and policy vary. Leading, all together, to the most diverse domain IOD benchmark, providing a complex challenge for any method. Table~\ref{tab:data_table} lists the tasks, their names, classes, and brief descriptions.

\begin{table*}[htbp]
\scriptsize
\centering
\caption{Description for D-RICO benchmark that consists of 15 tasks from 14 different datasets\textit{ incorporating} variations in multiple different aspects.}
\label{tab:data_table}
\begin{tabular}{cllll}
\toprule
\textbf{Task Number} & \textbf{Task Name}       & \textbf{Dataset}                                               & \textbf{Classes}               & \textbf{Short Description}     \\ \midrule
1     & \textit{daytime}               & nuImages~\cite{caesar_nuscenes_2020}                  & person, bicycle, vehicle                  &
urban, daylight, real-world, vehicle-mounted, Singapore
\\

2     & \textit{thermal}               & Teledyne FLIR~\cite{teledyne_flir_free_2024}        & person, bicycle, vehicle                     &
thermal, urban, varying lighting, weather conditions
\\

3     & \textit{fisheye fix}           & FishEye8K~\cite{gochoo_fisheye8k_2023}                & person, vehicle             &
fisheye, daytime, urban traffic, Taiwan, wide-angle, multi-camera        
\\

4     & \textit{drone}                 & VisDrone~\cite{zhu_detection_2022}                    & person, bicycle, vehicle                     &
drone, urban and rural, variable density, different lighting, 14 cities    
\\

5     & \textit{simulation}            & SHIFT~\cite{sun_shift_2022}                           & person, bicycle, vehicle                    &
synthetic, urban driving, CARLA, daytime, clear weather       
\\

6     & \textit{fisheye car}           & WoodScape~\cite{yogamani_woodscape_2021}              & person, vehicle             &
fisheye, vehicle-mounted, driving perspectives, multiple positions    
\\

7     & \textit{RGB + thermal fusion}        & SMOD~\cite{chen_amfd_2024}                            & person, bicycle, vehicle                    &
RGB-thermal fusion using IFCNN~\cite{zhang_ifcnn_2020}
\\

8     & \textit{video game}            & Sim10k~\cite{johnson-roberson_driving_2017}           & vehicle                     &
synthetic, urban, GTA V, diverse driving scenarios      
\\

9     & \textit{nighttime}             & BDD100K~\cite{yu_bdd100k_2020}                        & person, bicycle, vehicle                    &
urban, nighttime, perception challenge, street lighting
\\

10    & \textit{fisheye indoor}        & LOAF~\cite{yang_large-scale_2023}                     & person                      &
fisheye, indoor, overhead, 360° view, surveillance
\\

11    & \textit{gated}                 & DENSE~\cite{bijelic_seeing_2020}                      & person, vehicle             &
gated, urban, various conditions, depth-enhanced imaging      
\\

12    & \textit{photoreal. simulation}      & Synscapes~\cite{wrenninge_synscapes_2018}             & person, vehicle             &
photorealistic, synthetic, urban, physically based rendering     
\\

13    & \textit{thermal fisheye indoor}          & TIMo~\cite{schneider_timodataset_2022}                & person                      &
thermal fisheye, indoor, human actions, multiple perspectives
\\

14    & \textit{inclement}             & DENSE~\cite{bijelic_seeing_2020}                      & person, vehicle             &
fog, snow, rain, adverse weather        
\\

15    & \textit{event camera}          & DSEC~\cite{gehrig_low-latency_2024, gehrig_dsec_2021} & person, bicycle, vehicle                    &
event-based, driving, varied lighting, RGB overlay

\\\bottomrule
\end{tabular}
\end{table*}

\subsection{Evaluation Metrics}\label{sec:benchmark:metrics}
To assess IL performance, we adopt widely used metrics~\cite{anonymous_rico_2025, wang_comprehensive_2024, chaudhry_riemannian_2018}, using mean Average Precision ($\mAP$) as the primary evaluation criterion~\cite{menezes_continual_2022}. Our evaluation focuses on three aspects:

\begin{enumerate}
    \item \textbf{Overall effectiveness.}
    We measure aggregate performance with the average $\mAP$, denoted as $\overline{\mAP}$. Let $\mAP_{k,j}$ represent the $\mAP$ achieved on test set $\mathcal{D}_j$ of task $\mathcal{T}_j$ after completing training on task $\mathcal{T}_k$ (where $j \leq k$). The cumulative performance after task $k$ is defined as:
    \begin{equation}
        \overline{\mAP}_k = \frac{1}{k} \sum_{j=1}^{k} \mAP_{k,j},
    \end{equation}
    where larger values indicate better retention and generalization across tasks.

    \item \textbf{Retention and forgetting.}
    We evaluate memory stability via the \emph{forgetting measure} (FM), which captures the decline in a model’s performance on earlier tasks. After training on task $k$, the forgetting metric is computed as:
    \begin{equation}
        \mathrm{FM}_k = \frac{1}{k-1} \sum_{j=1}^{k-1} \max_{1\le l\le k-1} \left(\mAP_{l,j} - \mAP_{k,j} \right).
    \end{equation}
    A higher FM value reflects increased forgetting, while negative values suggest performance gains on prior tasks.

    \item \textbf{Adaptability and transfer.}
    A model's ability to learn new tasks effectively is characterized by two complementary metrics:
    \begin{enumerate}
        \item \emph{Forward transfer} (FWT) quantifies how previously acquired knowledge benefits learning a new task. It is calculated as:
        \begin{equation}
            \mathrm{FWT}_k = \frac{1}{k-1} \sum_{j=2}^{k} \left(\mAP_{j,j} - \mAP'_{j}\right),
        \end{equation}
        where $\mAP'_{j}$ denotes the performance of an independently trained model on task $\mathcal{T}_j$. Positive FWT indicates improved learning due to prior experience.
        
        \item \emph{Intransigence} (IM) assesses the difficulty in learning new tasks relative to a jointly trained model. It is defined as:
        \begin{equation}
            \mathrm{IM}_k = \frac{1}{k} \sum_{j=1}^{k} \left( \mAP_{j,j} - \mAP^*_j\right),
        \end{equation}
        where $\mAP^*_j$ corresponds to the $\mAP$ obtained from a model trained on all task data $\cup_{j=1}^{T} \mathcal{D}_j$ simultaneously. A positive IM implies greater adaptability than joint training.
    \end{enumerate}
\end{enumerate}

At the conclusion of all $T$ tasks, we denote the final metric values as $\overline{\mAP}=\overline{\mAP}_T$, $\mathrm{FM}=\mathrm{FM}_T$, $\mathrm{FWT}=\mathrm{FWT}_T$, and $\mathrm{IM}=\mathrm{IM}_T$.

The overarching objective is for IL models to surpass both standalone and joint models by leveraging inter-task transfer, ideally satisfying $\overline{\mAP}>\frac{1}{T} \sum_{j=1}^{T} \mAP_{j}'$, which necessitates high adaptability and minimal forgetting.

\subsection{Results}
We first present the main results of three prompt-based methods and reference baselines on the D-RICO benchmark, followed by additional analyses on initialization and prompt length .We choose a diverse subset of five tasks, \ie $[1, 2, 3, 11, 15]$, from the 15 D-RICO tasks for the main results and all 15 tasks in the subsequent experiments.

\subsubsection{Main Results}
\begin{figure*}[tbh]
    \centering
    \includegraphics[width=0.9\linewidth]{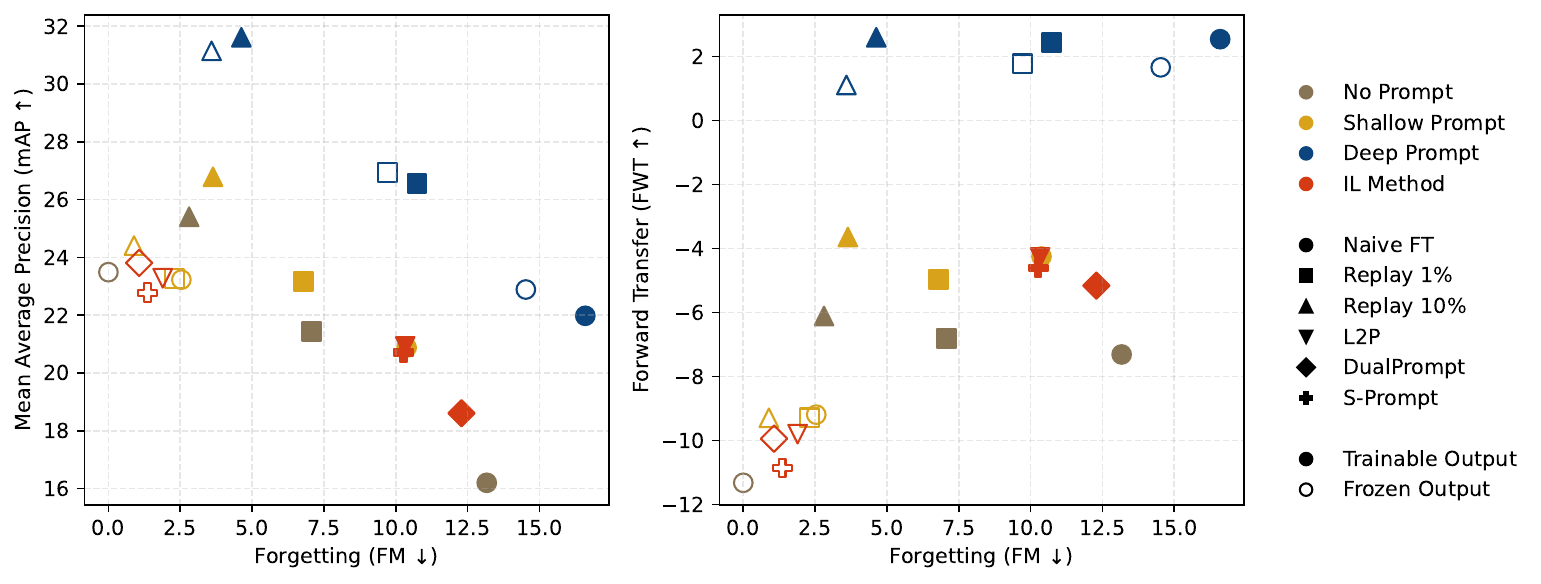}
    \caption{Incremental learning results on D-RICO benchmark. The left figure shows overall performance $\overline{\bm{\mAP}}$ versus the forgetting (FM) and the right shows plasticity (FWT) versus FM. The three prompt-based IL methods are far from the optimal of high plasticity and low forgetting (upper left corner).}
    \label{fig:main-results}
\end{figure*}

The main results on the D-RICO benchmark are shown in Table~\ref{tabel:main-results} and Figure~\ref{fig:main-results}. Among the three prompt-based IL methods, L2P achieves the highest performance when the output layer is not frozen, while DualPrompt slightly outperforms the other two methods when the output layer is fixed after the first task. Regarding forgetting, DualPrompt is also the lowest.

The three prompt-based IL methods perform similarly to Naïve FT and lag substantially behind replay at both 1\% and 10\%. In Figure~\ref{fig:main-results}, this becomes more obvious where they show high forgetting while having mediocre overall performance and plasticity. However, as all three methods do not employ deep prompting, in a fair comparison to shallow prompting, they achieve a similar performance to replay 1\%, though Naïve FT is also close to that.

Fixing the output layer generally benefits all IL settings except for the 10\% replay scenario. Specifically, weaker methods such as Naïve FT, replay 1\%, L2P, DualPrompt, and S-Prompt all benefit from reduced model plasticity, as their counterparts with non-fixed output layers exhibit lower performance in terms of $\overline{\bm{\mAP}}$ and FM. However, strong regularization via 10\% replay benefits from increased plasticity, enabling it to surpass individually trained models in both shallow and deep prompt scenarios.

Overall, deep prompting consistently outperforms shallow prompting regarding $\overline{\bm{\mAP}}$ and FWT, although shallow prompting demonstrates lower FM. The two plasticity metrics (FWT and IM) show an increase in model adaptability. For Naïve FT, deep and shallow prompting yield similar $\overline{\bm{\mAP}}$, highlighting a trade-off between stability (FM) and plasticity (FWT). Employing prompts generally outperforms the no-prompt condition. However, when the output layer is fixed (i.e., no further learning occurs after the initial task), the $\overline{\bm{\mAP}}$ performance of the no-prompt condition becomes similar to the three prompt-based methods. This further illustrates that these standard methods are not sufficiently competitive on this challenging benchmark.

\subsubsection{Prompt Length}\label{subsec:prompt-length}
It is expected that different tasks require varying prompt lengths depending on their diversity. To illustrate this, we train each task in the D-RICO benchmark with different prompt lengths (1, 5, 10, 25, 100, 250, 500) to identify the optimal length for each. The results shown in Table~\ref{tab:prompt-length} confirms this across three different prompting techniques. It is evident that some tasks perform well with a single prompt, while others require up to 500. Choosing the best prompt length for each task slightly increases the final mAP. However, the difference compared to the next-best fixed prompt length is minimal.

\begin{figure}
    \centering
    \includegraphics[width=1\linewidth]{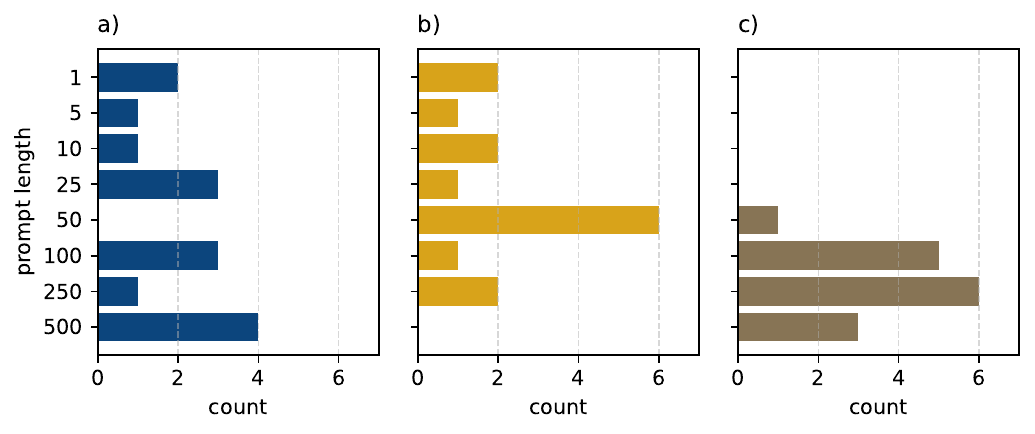}
    \caption{Count for how often a prompt length led to the best mAP in each of the three prompting categories. Plot a) shows result for shallow prompting where the prompt is removed after the first layer, b) shallow with keeping the prompt and c) deep prompt. The results demonstrate that larger prompt length work well for deep prompting, and shallow prompting requires a bit less.}
    \label{fig:prompt-length-winnerx}
\end{figure}

Figure~\ref{fig:prompt-length-winnerx} shows a histogram of how often a prompt length yields the best outcome. When there's a tie, the shorter length is selected because it's more computationally efficient and thus preferred. It is clear that the optimal prompt length depends on the prompting style (shallow versus deep and remove versus keep prompt) and the task. Generally, deep prompting can better utilize longer prompt lengths compared to shallow prompting. If the prompt is removed after the first layer, i.e., it only influences the first MSA, longer prompt lengths work better than when the prompt is kept in the model.

\begin{table*}[tb]
    \centering
    \footnotesize
    \setlength{\tabcolsep}{1.5pt}

    \caption{Analysis of optimal prompt length across three prompt techniques for each D-RICO task shows that not all prompt configurations outperform training without prompts. Generally, shallow prompts offer only marginal gains. Selecting the optimal prompt length per task yields the best average performance, though the improvement over a fixed prompt length is minimal.}
    \label{tab:prompt-length}
   \begin{tabular}{c|c|ccccccccc|ccccccccc|ccccccccc}
        \toprule
 & & \multicolumn{9}{c|}{Shallow Prompt (Remove Prompt)} & \multicolumn{9}{c|}{Shallow Prompt (Keep Prompt)} & \multicolumn{9}{c}{Deep Prompt} \\
        \cmidrule(lr){3-11} \cmidrule(lr){12-20} \cmidrule(lr){21-29}
Task & 0 & 1 & 5 & 10 & 25 & 50 & 100 & 250 & 500 & Best & 1 & 5 & 10 & 25 & 50 & 100 & 250 & 500 & Best & 1 & 5 & 10 & 25 & 50 & 100 & 250 & 500 & Best \\
        \midrule
        1 & 41.2 & 41.0 & 41.3 & 41.2 & \textbf{41.6} & 41.2 & 41.3 & 41.2 & 41.4 & 41.6 & 41.2 & \textbf{41.4} & 41.2 & 41.2 & 41.1 & 41.0 & 41.0 & 40.5 & 41.4 & 42.5 & 43.4 & 43.6 & 43.9 & 44.4 & 43.9 & \textbf{44.5} & 44.5 & 44.5 \\
        2 & 33.2 & 33.2 & 34.1 & 34.1 & 34.2 & 34.6 & 34.5 & 34.8 & \textbf{35.3} & 35.3 & 33.4 & 33.4 & 34.2 & 34.8 & 34.7 & 35.0 & \textbf{35.3} & 34.6 & 35.3 & 37.7 & 38.5 & 39.1 & 39.3 & 39.9 & \textbf{40.3} & 39.8 & 39.9 & 40.3 \\
        3 & 20.3 & 19.6 & 19.7 & 20.0 & 20.0 & 20.5 & 20.5 & 20.3 & \textbf{20.6} & 20.6 & 20.0 & 20.1 & 20.3 & 20.3 & 20.4 & 20.3 & \textbf{20.7} & 20.6 & 20.7 & 23.1 & 24.0 & 24.2 & 24.8 & \textbf{25.0} & 24.8 & 25.0 & 24.4 & 25.0 \\
        4 & 18.7 & 18.7 & 18.6 & 18.6 & 18.7 & 18.6 & 18.6 & \textbf{18.8} & 18.7 & 18.8 & 18.7 & 18.8 & 18.8 & 18.9 & 18.9 & \textbf{19.0} & 18.8 & 18.8 & 19.0 & 20.2 & 20.4 & 20.8 & 20.9 & 21.1 & \textbf{21.2} & 21.1 & 21.2 & 21.2 \\
        5 & 30.7 & \textbf{30.2} & 29.8 & 29.7 & 29.9 & 29.6 & 29.6 & 29.1 & 29.7 & 30.2 & 30.1 & 30.1 & \textbf{30.3} & 30.0 & 29.8 & 29.7 & 29.6 & 29.3 & 30.3 & 31.1 & 31.6 & 32.0 & 32.3 & 32.5 & 32.6 & \textbf{32.7} & 32.5 & 32.7 \\
        6 & 39.9 & 39.8 & 39.7 & 39.7 & \textbf{40.0} & 39.8 & 39.7 & 39.9 & 39.9 & 40.0 & 39.8 & 39.9 & 39.9 & 39.9 & \textbf{40.1} & 40.0 & 39.9 & 39.3 & 40.1 & 41.1 & 41.7 & 41.9 & 42.3 & 42.0 & 42.3 & 42.2 & \textbf{42.4} & 42.4 \\
        7 & 44.9 & 44.6 & 44.6 & \textbf{45.0} & 44.9 & 45.0 & 44.9 & 45.0 & 44.7 & 45.0 & 44.9 & 44.8 & \textbf{45.0} & 44.8 & 45.0 & 44.2 & 44.2 & 43.4 & 45.0 & 47.1 & 49.8 & 50.0 & 50.4 & 50.4 & 50.4 & \textbf{50.8} & 50.3 & 50.8 \\
        8 & 44.8 & 44.9 & 45.0 & 45.0 & 45.0 & 44.8 & \textbf{45.1} & 44.7 & 44.7 & 45.1 & \textbf{45.2} & 45.0 & 45.1 & 45.2 & 45.1 & 44.8 & 44.6 & 44.4 & 45.2 & 45.8 & 46.1 & 46.5 & 46.6 & 46.3 & 46.5 & 46.1 & \textbf{46.8} & 46.8 \\
        9 & 20.5 & 21.0 & 21.2 & 20.7 & 21.5 & 20.8 & \textbf{21.8} & 21.5 & 21.7 & 21.8 & 20.4 & 21.3 & 21.1 & 21.3 & \textbf{21.5} & 21.4 & 20.9 & 21.1 & 21.5 & 21.3 & 22.2 & 22.3 & 22.9 & 22.7 & 23.1 & \textbf{23.4} & 22.7 & 23.4 \\
        10 & 37.3 & 37.7 & 37.9 & 37.4 & \textbf{38.2} & 37.6 & 37.5 & 38.0 & 37.4 & 38.2 & 37.4 & 37.9 & 37.9 & 37.8 & \textbf{38.0} & 37.2 & 36.9 & 36.9 & 38.0 & 43.1 & 43.9 & 43.4 & 43.2 & 44.3 & \textbf{44.4} & 44.4 & 43.4 & 44.4 \\
        11 & 27.7 & 27.8 & \textbf{28.2} & 27.5 & 27.8 & 27.8 & 27.4 & 27.7 & 27.5 & 28.2 & \textbf{28.1} & 27.2 & 27.3 & 26.6 & 27.0 & 24.7 & 27.4 & 21.0 & 28.1 & 30.7 & 31.5 & 32.2 & 32.5 & 32.9 & \textbf{33.5} & 33.2 & 33.2 & 33.5 \\
        12 & 24.6 & \textbf{24.1} & 24.0 & 23.8 & 24.0 & 24.1 & 24.0 & 23.9 & 24.0 & 24.1 & 24.1 & 24.2 & 24.1 & 24.3 & \textbf{24.5} & 24.5 & 24.4 & 24.3 & 24.5 & 25.4 & 25.9 & 26.1 & 26.2 & 26.4 & 26.5 & 26.6 & \textbf{26.7} & 26.7 \\
        13 & 69.6 & 72.1 & 71.4 & 70.9 & 72.4 & 72.3 & 72.6 & 72.0 & \textbf{73.6} & 73.6 & 69.8 & 70.5 & 72.6 & 72.9 & \textbf{73.0} & 73.0 & 71.1 & 71.8 & 73.0 & 80.2 & 82.6 & 82.7 & 83.4 & 83.6 & \textbf{84.7} & 84.3 & 84.0 & 84.7 \\
        14 & 48.5 & 48.6 & 48.6 & 48.5 & 48.6 & 48.5 & 48.6 & 48.5 & \textbf{48.7} & 48.7 & 48.7 & 48.6 & 48.6 & \textbf{48.7} & 48.6 & 48.3 & 48.4 & 47.9 & 48.7 & 49.4 & 50.0 & 50.0 & 51.0 & 50.9 & 50.9 & \textbf{51.1} & 51.1 & 51.1 \\
        15 & 12.2 & 17.1 & 20.1 & 18.4 & 19.2 & 19.3 & \textbf{21.3} & 18.9 & 20.4 & 21.3 & 15.8 & 18.8 & 18.3 & 21.0 & \textbf{21.7} & 18.7 & 20.0 & 20.1 & 21.7 & 20.3 & 23.5 & 21.3 & 23.4 & 23.2 & 23.1 & \textbf{25.4} & 23.2 & 25.4 \\
        \midrule
        Mean & 34.3 & 34.7 & 34.9 & 34.7 & 35.1 & 35.0 & 35.2 & 35.0 & 35.2 & \textbf{35.5} & 34.5 & 34.8 & 35.0 & 35.2 & 35.3 & 34.8 & 34.9 & 34.3 & \textbf{35.5} & 37.3 & 38.3 & 38.4 & 38.9 & 39.0 & 39.1 & 39.4 & 39.1 & \textbf{39.5} \\
        \bottomrule
    \end{tabular}
\end{table*}

\subsubsection{Prompt Initialization}\label{subsec:prompt-init}
Previous works on visual prompt methods for IL used uniform prompt initialization with random values between -1 and 1. We noticed in preliminary experiments that we can achieve better results with smaller intervals. To study this further, we run experiments for the initializations values $\mathrm{init} \in \{10^{-6}, 10^{-5}, 10^{-4}, 10^{-3}, 10^{-2}, 10^{-1}, 1\}$, meaning prompt initialization with uniform random values in the interval $[-\mathrm{init}, \mathrm{init}]$, for different configurations. For the different configurations we use prompt lengths $L_P\in \{1, 5, 10, 25, 50, 100, 250, 500\}$ and injection layers $\mathrm{inject} \in \{\allowbreak [0],\allowbreak [0, 1, 2, 3],\allowbreak [0, 1, 2, 3, 4, 5, 6],\allowbreak [0, 1, 2, 3, 4, 5, 6, 7, 8, 9, 10, 11, 12],\allowbreak [7, 8, 9, 10, 11, 12]\}
$. We only train on task 4. 

\begin{figure}
    \centering
    \includegraphics[width=1\linewidth]{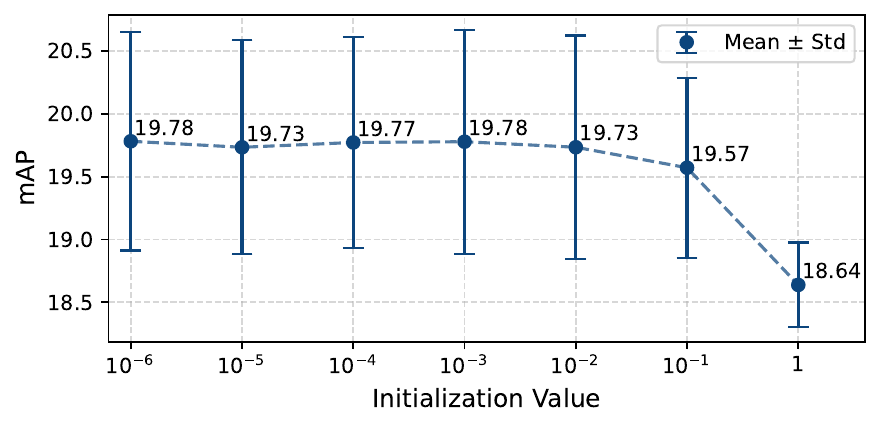}
    \caption{Results of different prompt initialization intervals $[-\mathrm{init}, \mathrm{init}]$ averaged over various prompt lengths and injection layers for task 4 from D-RICO. For lower values, the results stabilize and are better than for larger intervals.}
    \label{fig:prompt-init}
\end{figure}

Figure~\ref{fig:prompt-init} displays the average results for various initialization configurations. It is evident that the commonly used interval $[-1, 1]$ does not produce the best outcomes. Below $10^{-2}$, results level off, indicating that for low uniform initialization values, the specific value chosen has little effect on the outcome. The standard deviation remains similar across all low initialization values and decreases slightly towards 1.

\section{Discussion}
In this section, we collectively summarize and discuss these findings, with key takeaways provided in the text box.

The results of DualPrompt in Table~\ref{tabel:main-results} demonstrate the general feasibility of employing prompt-based IL methods for domain IOD. However, all three tested methods underperform compared to randomly replaying data, highlighting the necessity for more advanced prompt-based methods. A wide variety of methods developed for classification could be explored in future work.

The study of the reference baselines indicates that deep prompting significantly outperforms shallow prompting. Thus, future research should focus on deep prompts to enhance overall performance and increase plasticity.

As observed, replaying just 1\% of data from previous tasks represents a simple yet robust baseline. Expanding the replay buffer to 10\% further reduces forgetting and improves overall performance. 

Naive FT, replay 1\%, L2P, DualPrompt, and S-Prompt all benefit from fixing the output layer after the initial task. Strong regularization, as employed in the 10\% replay scenario, further improves performance due to increased available plasticity. Therefore, future methods should also consider adaptations at the output layer, as modifications solely in the feature space are insufficient for achieving optimal IL performance.

Determining the optimal prompt length is not straightforward, as it varies depending on the specific prompting technique and task. While selecting an individual length per task provides minor advantages, the benefits currently do not justify the complexity and additional hyperparameter tuning required. However, this aspect could become relevant in future benchmarks or practical applications. Generally, object detection requires longer prompts compared to classification tasks.

In contrast, the influence of prompt initialization on performance is significant. Results presented in Figure~\ref{fig:prompt-init} suggest that initializing with smaller random values from the uniform interval $[-10^{-2}, 10^{-2}]$ yields superior and more stable performance. This differs from prompt-based classification methods, where random initialization typically occurs within a larger interval, such as $[-1, 1]$. A better initialization scheme can notably improve results.

Future investigations and experiments should assess the performance of class incremental learning (CIL) and few-shot IL for DIL and CIL. CIL is particularly challenging, as prompt-based methods only modify the backbone, necessitating an additional mechanism to address forgetting in an expanding head. Since the studied D-RICO benchmark surpasses existing benchmarks in diversity, we anticipate that the results will apply to these less diverse benchmarks.

\begin{tcolorbox}[
colback=keytakeaways,
colframe=white,
arc=4mm,
boxrule=0pt,
left=2mm,
right=2mm,
top=1mm,
bottom=1mm,
width=\linewidth
]
\small
\centering \textbf{Key Takeaways}

\vspace{1mm}

\raggedright
\textbf{Feasibility.} Visual-prompt methods studied here provide minimal help in mitigating catastrophic forgetting for domain incremental learning.\\
\vspace{1mm}
\textbf{Best Method.} DualPrompt~\cite{wang_dualprompt_2022} performs best among the tested prompt-based approaches.\\
\vspace{1mm}
\textbf{Deep vs. Shallow.} Deep prompts significantly outperform shallow prompts.\\
\vspace{1mm}
\textbf{Output Layer.} Unfreezing the output layer, paired with strong regularization, enhances performance by increasing plasticity. Weak methods profit from freezing the output layer after the first task.\\
\vspace{1mm}
\textbf{Replay.} Replaying even 1\% of previous data surpasses prompt-based IL methods, especially in combination with deep prompting.\\
\vspace{1mm}
\textbf{Prompt Length.} Deep prompting benefits more from longer prompts. Using an individual prompt length for all tasks yields only minimal improvement in the results.\\
\vspace{1mm}
\textbf{Prompt Initialization.} Random initialization values for the prompts within $[-10^{-2}, 10^{-2}]$ or narrower yield optimal performance.
\end{tcolorbox}

\section{Conclusion}
This work presents the first comprehensive analysis of prompt-based IL methods for object detection. We evaluated three classification-derived approaches—L2P~\cite{du_learning_2022}, DualPrompt~\cite{wang_dualprompt_2022}, and S-Prompt~\cite{wang_s-prompts_2022}—against a range of strong reference baselines on the challenging D-RICO benchmark~\cite{anonymous_rico_2025}. Our findings confirm the general feasibility of applying prompt-based IL to object detection, with DualPrompt achieving the highest performance among the prompt-based methods. However, all evaluated methods are still outperformed by simple replay-based strategies, underscoring the need for further innovation in prompt design and learning mechanisms. We believe that our empirical insights will serve as valuable guidance for advancing prompt-based IL methods in object detection.

{
    \small
    \bibliographystyle{ieeenat_fullname}
    \bibliography{first-paper}
}

\end{document}